# NON-INVASIVE CALIBRATION OF A STEWART PLATFORM BY PHOTOGRAMMETRY

**Sourabh Karmakar**
PhD Student
Mechanical Engineering
Clemson University
Clemson, SC
sourabh.karmakar@gmail.com

**Cameron J. Turner**[1]
Associate Professor
Mechanical Engineering
Clemson University
Clemson, SC
cturne9@clemson.edu

## ACKNOWLEDGEMENTS

The authors would like to acknowledge the support of Clemson University. All statements within are those of the authors and may or may not represent the views of these institutions.

---

[1] Corresponding Author

**ABSTRACT**

*Accurate calibration of a Stewart platform is important for their precise and efficient operation. However, the calibration of these platforms using forward kinematics is a challenge for researchers because forward kinematics normally generates multiple feasible and unfeasible solutions for any pose of the moving platform. The complex kinematic relations among the six actuator paths connecting the fixed base to the moving platform further compound the difficulty in establishing a straightforward and efficient calibration method. The authors developed a new forward kinematics-based calibration method using Denavit-Hartenberg (DH) convention and used the Stewart platform "Tiger 66.1" developed in their lab for experimenting with the photogrammetry-based calibration strategies described in this paper. This system became operational upon completion of construction, marking its inaugural use. The authors used their calibration model for estimating the errors in the system and adopted three compensation options or strategies as per Least Square method to improve the accuracy of the system. These strategies leveraged a high-resolution digital camera and off-the-shelf software to capture the poses of the moving platform's center. This process is non-invasive and does not need any additional equipment to be attached to the hexapod or any alteration of the hexapod hardware. This photogrammetry-based calibration process involves multiple high-resolution images from different angles to measure the position and orientation of the platform center in the three-dimensional space. The Target poses and Actual poses are then compared, and the error compensations are estimated using the Least-Squared methods to calculate the Predicted poses. Results from each of the three compensation approaches demonstrated noticeable enhancements in platform pose accuracies, suggesting room for further improvements. Given that "Tiger 66.1" is based on the general Stewart Platform structure, the proposed calibration method holds promise for extension to machines*

*operating on similar principles where non-invasive calibration is desirable. This study contributes to advancing the field of Stewart platform calibration, paving the way for more precise and efficient applications in various domains.*

**Keywords:** Calibration, Stewart Platform, Photogrammetry, Least-Square method, Inverse Kinematics, Forward Kinematics.

## 1. Introduction

Parallel Kinematic Machine (PKMs) constitute a pivotal domain within robotics. Any PKM is characterized by a fixed base and a moving platform. The base and platform are connected by multiple parallel actuators, and the number of actuators can vary between 3 to 6. The actuators are used for controlling position and orientation of the platform. The machines with six actuators are called Stewart Platform, a Stewart-Gough Platform, a Gough-Stewart Platform or more commonly, a hexapod. Hexapod platforms stands out as one of the most prominent and widely adopted parallel kinematic machines [1]. The six actuators add six degrees of freedom at the center of the moving platform [2]. To articulate the dynamics of the moving platform central point, a cartesian coordinate-frame is attached to the platform center. The configuration of the platform center is specified by three translatory or linear displacement along $x$, $y$, $z$ axes and three rotations about the same $x$, $y$, $z$ axes [3] from a reference position called the *home pose*. The position and orientation of the platform is dependent on the actuators and joints connecting the fixed base with the moving platform. The joints can be spherical (S) or universal (U) while the actuators provide linear motions through prismatic (P) joints. The hexapod, named "Tiger 66.1", used by the authors has six actuators acting as six prismatic (P) actuated joints, each of them connected to the fixed base with a universal (U) joint at one end and the other end with a spherical joint (S) connected to the moving platform. The combination of joints defines the designation of the hexapod. This configuration classifies Tiger 66.1 as a 6-UPS (total 6 DOF through 1 Universal, 1 Prismatic, and 1 Spherical joint) Parallel Kinematic Machine [4]. This hexapod has been built by the authors in their lab and is used in this study to experiment with photogrammetry-based calibration.

Calibrations of a Stewart-Gough platform has attracted research interests in the last couple of decades [5]. Various methods were adopted to make the calibration processes simple and straightforward, using both forward and inverse kinematics. The forward kinematics for hexapod are complex and difficult due to its non-linear kinematics equations and multiple solutions [6]. Moreover, the use of additional equipment and / or modification of the hexapod hardware makes the calibration process more complicated. Various pieces of equipment were used to conduct the calibration procedures. Zhuang *et al.* [7] used commercial electronic theodolite for the calibration of their hexapod platform. Ryu J., and Rauf A. [8] imposed constraint motion on the end-effector by fixing the length of one of the six actuators and through inverse kinematics. In another research effort, Großmann *et al.* [9] used a simple and robust double-ball-bar (DBB) for measurements from a continuously moving hexapod platform under six degrees of freedom. Liu *et al.* [10] innovatively adopted self-calibration, incorporating a three-dimensional laser tracker and a genetic algorithm into their calibration method, which involved both simulated and real measurements.

Digital cameras have assumed a significant role in hexapod calibration research. Daney et al. [11] harnessed a Sony digital video camera for measuring the joint positions and leg lengths on their hexapod named "Table of Stewart". They seamlessly integrated inverse kinematics with data obtained from the digital camera to implement their calibration methodology. A omni-directional camera was employed by Dallej et al. [12] in their lab to measure the positions and orientations of the actuators in their hexapod, leveraging inverse kinematics for calibration. A high resolution digital camera was used by Nategh et al. [13] for the calibration process. The camera was used to capture images of the moving platform in a vertically downward direction. They developed MATLAB code to extract the platform's pose across various positions and orientations from these images.

With the continuous evolution of digital cameras and the increasing availability of software and hardware support, the integration of image processing into calibration methods is increasing. Notably, while photogrammetry has witnessed substantial adoption across various fields, its application in hexapod calibration remains relatively limited. This present study exclusively relied on photogrammetry. A high-resolution digital camera Nikon D3200 with AF-S DX Nikkor 18-105mm lens has been used. For image processing, photogrammetry software "Photomodeler 2023" and PTC Creo 9.0.5 student edition have been used.

The error model formulated by the authors in this study exclusively incorporates the error quantified in the pose of the moving platform; no additional errors were measured or considered for compensation in this study. In this model, the contributions of any other errors like joint errors, errors originating from any other sources of the machine were omitted, they are unnecessary in this approach at this stage. So, the hexapod platform has not been equipped with any additional sensors for measuring errors other than only the pose error which has been measured through photogrammetry. The methods proposed in this study utilized inverse kinematics to derive forward kinematics solutions for platform poses. From the forward kinematic solutions, the DH parameters for each actuator path were calculated using modified Denavit-Hartenberg (DH) convention. Although the DH convention is the most popular method for forward kinematics in serial manipulators, its application remains very limited in parallel manipulators due to the closed loop nature of PKMs [14] and generation of multiple feasible and unfeasible solutions for any pose of parallel robots. Here modified DH convention was adopted because of its simplicity and straight-forward nature of implementation. This was possible by the new algorithm [15] developed by the authors to find unique, feasible forward kinematic solution for any pose in PKMs. The DH parameters obtained from the target pose and corresponding actual pose were compared and analyzed to develop the calibrated predicted pose for the target configuration. As initial experiments, all the error compensations were calculated with Least-Squared methods.

The subsequent sections of this paper are organized as follows: The introduction has been presented in this initial section. The second part illustrates the calibration methodology used by the authors for calibrating their "Tiger 66.1" hexapod. The succeeding section explains the experimental setup and data collection methods used. In Section 4, the data collected has been analyzed and findings were documented. Section 5 engages in discussions pertaining to the analyzed data. The concluding remarks were shared in the last section.

## 2. Calibration Methodology

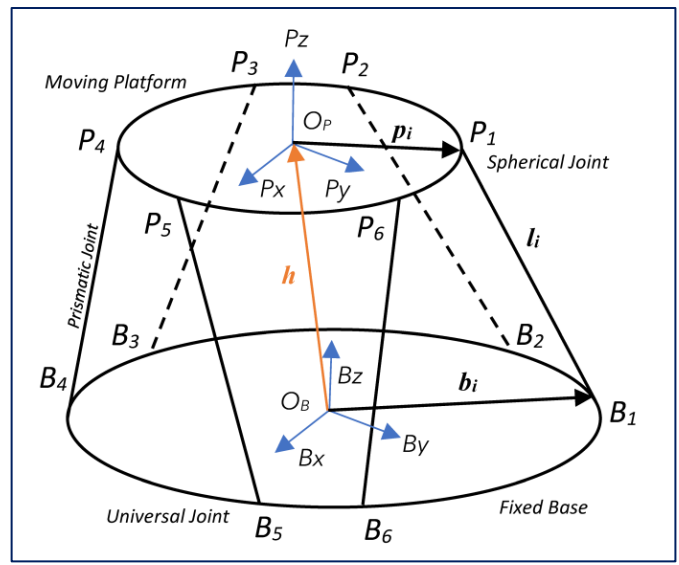

Figure 1: A typical hexapod configuration

In Figure 1, the typical sketch of a hexapod platform [15] is illustrated. $O_B$ is the center of the fixed base and $O_P$ is the center of the moving platform. Two cartesian coordinate frames $O_BB_xB_yB_z$ and $O_PP_xP_yP_z$ are attached with these center points. The cartesian coordinate frame with $O_B$ as origin has *x*, *y*, *z* axes denoted by $B_x$, $B_y$ and $B_z$ respectively and for moving frame with origin $O_P$ has $P_x$, $P_y$ and $P_z$ axes to indicate *x*, *y*, *z* axes respectively. $B_z$ and $P_z$ are the vertical axes of the respective coordinate frames. The configuration of the moving platform is defined by the position and orientation of coordinate frame attached at its center with respect to base coordinate frame by 6 parameters: distance along *x*, *y*, *z* axes and rotations about the same three axes.

Figure 2 shows the 3D CAD model of Tiger 66.1 and the actual platform used for the experiments. Tiger 66.1 has been developed for characterizing additively manufactured materials under complex loading conditions including tension, torsion, bending and combinations thereof[16].

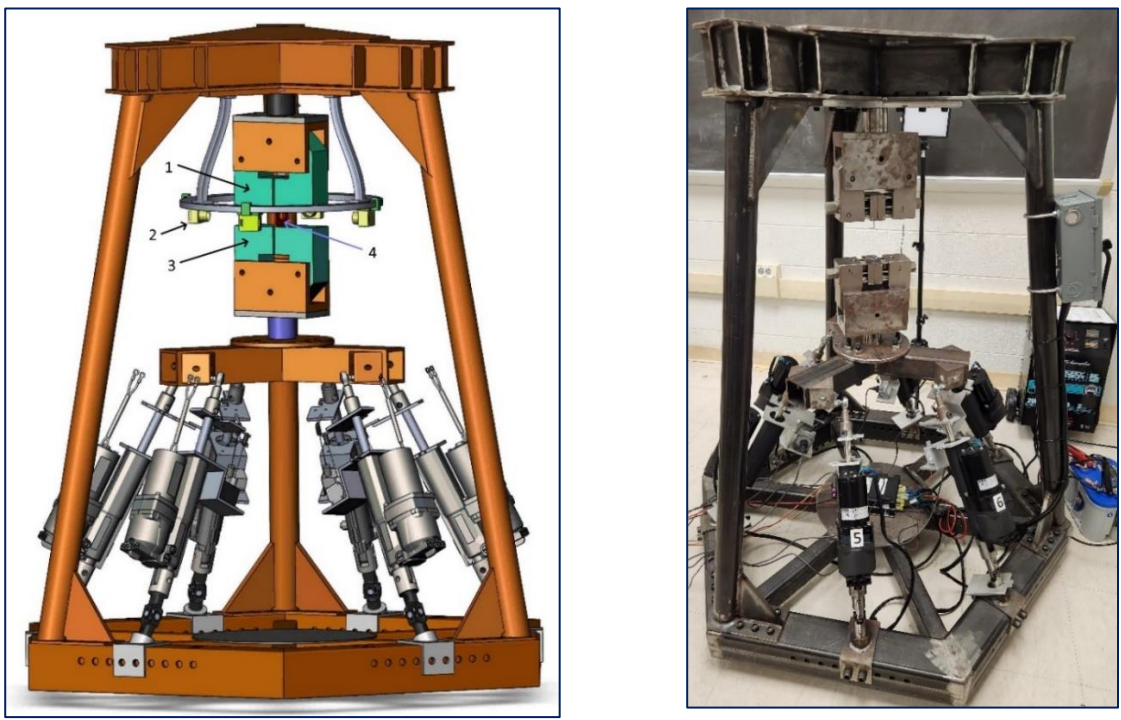

Figure 2: CAD model of hexapod Test Frame "Tiger 66.1" & the actual test frame

The configuration of a general Stewart platform has been modified to make the machine suitable for material testing while keeping the basic principle of Stewart platform unchanged. The fixed base has been extended through rigid structures to move the fixed coordinate frame to the top of the moving platform. This has been done to install a fixed gripper in a suitable position for the convenience of material characterization tests. The two green blocks as shown in the CAD model, on the upper part of the system are the grippers for holding the material specimen (4) to be tested. The upper gripper (1) is mounted on the fixed frame and the lower gripper (3) is fixed at the center of the hexapod moving platform. All motions and forces are applied on the test specimen by moving the lower gripper. Photogrammetry has been planned to use for measurements in this test process, so there are provisions to fix four digital cameras (2) to capture images from the test zone. As the manufactured platform is at the beginning stage of its development, these cameras have not yet been installed. Instead, all images were captured from different angles by using an external camera with a suitable field-of-view as mentioned previously.

To use photogrammetry, two new coordinate frames have been introduced in Tiger 66.1 for all measurements and calculations: one at the center on the upper grip end plate and another at the center on the lower grip end plate. The new frame configurations have been shown in Figure 3. $O_{UG}$ is the center of the fixed upper grip end plate and the associated coordinate axes are $UG_x$, $UG_y$ and $UG_z$ for *x*, *y*, *z* axes respectively. Similarly, $O_{LG}$ is the center of the movable lower grip end plate and the associated coordinate axes are $LG_x$, $LG_y$ and $LG_z$ corresponding to *x*, *y*, *z* axes respectively. The distance between $O_{UG}$ and $O_B$ are fixed and known and the distance *fd* between $O_{LG}$ and $O_P$ are also fixed and known. The distance *gd* between the fixed grip center $O_{UG}$ and moving grip center $O_{LG}$ will change during the operation of Tiger 66.1. The relationships between these frames are easily established by spatial transformation matrices. In the '*home pose*' [17] all the *z*-axes remain vertical and colinear while the other axes remain parallel to each other and all *x*-axes or *y*-axes remain in one plane. In the home pose the length of all actuators is equal and known.

The lower grip center is moved with the moving platform by controlling the lengths of the actuators and the resultant motion generated at the lower grip center is a translation or rotation or a combination of both.

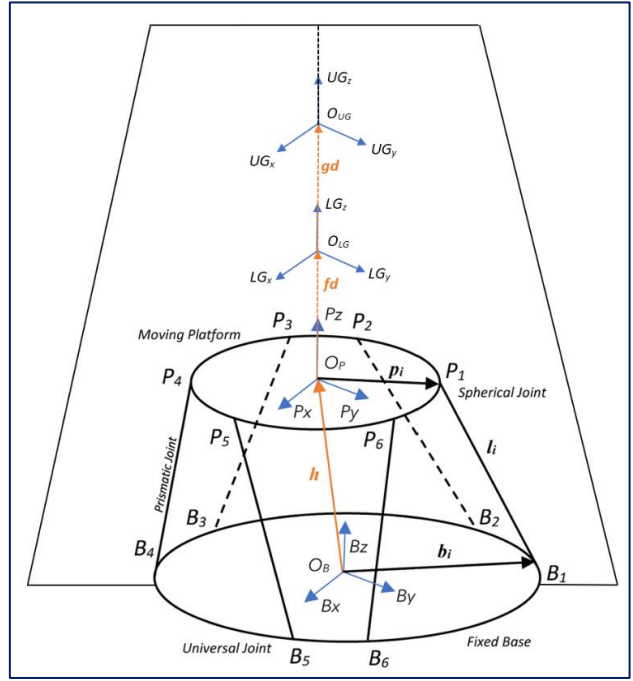

Figure 3: Tiger 66.1 coordinate frame configuration

The new position and orientation of the lower grip center depends on the values of roll, pitch, yaw, and the translation motion along $x$, $y$, $z$ axes as per the Euler angle representations [18]. These are values measured from $O_{LG}$ with respect to its home pose. In this investigation, the home pose has been specifically established at $gd$ = 50mm, in accordance with the other requisite home pose conditions. This $gd$ value is equal to the standard gauge length for tensile test of any material.

The rotations from the home pose are expressed by vector $\Phi$ and

$$\Phi = (\alpha\ \beta\ \gamma)^T \qquad (1)$$

where $\alpha$ (roll), $\beta$ (pitch), $\gamma$ (yaw) denotes the rotation angles about $x$, $y$ & $z$ axes respectively.

The translations are expressed by vector $\boldsymbol{d}$ where,

$$\boldsymbol{d} = (x \; y \; z)^T \tag{2}$$

and $x$, $y$ & $z$ are the translation values from home pose along $x$, $y$ and $z$ axes respectively.

For the calibration process, Tiger 66.1 always starts moving from the home pose to travel to each new pose. In the home pose all actuator lengths are equal. Once the new actuator lengths were calculated by inverse kinematics for each new pose, the platform controller is fed with new actuator lengths. Employing the digital camera, multiple photographs were taken for each pose from suitable angles and are processed through Photomodeler and ProEngineer Creo to get the new position of the lower grip center in the 3D space. The target pose values are subtracted from actual reached pose values to calculate the six error parameters in terms of $x$, $y$, $z$ positions and orientations. These error values were now used to calculate the error compensation for Tiger 66.1.

For a pose, if the target pose values = $(x \; y \; z \; \alpha \; \beta \; \gamma)^T$

and the real measurement shows the values = $(x' \; y' \; z' \; \alpha' \; \beta' \; \gamma')^T$,

then the error vector for a pose = $((x' - x) \; (y' - y) \; (z' - z) \; (\alpha' - \alpha) \; (\beta' - \beta) \; (\gamma' - \gamma))^T$

For $n$ number of poses, there will $n$ numbers of such error vectors.

The error values are used to calculate the correction values by Least-Square method. The correction values are then combined with target pose values to calculate the predicted poses.

The Least-Squares method [19] is a statistical method for fitting a line or curve to a set of data points. It minimizes the sum of the squared residuals, which are the distances between the data points and the fitted line or curve. The mathematical expression for the least squares method:

$$\text{Cost function} = \min \sum (y - f(x))^2 \tag{3}$$

where $y$ is the dependent variable, $x$ is the independent variable, $f(x)$ is the fitted line or curve, and $\sum$ is the sum of all the terms.

In other words, the least squares method finds the line or curve that minimizes the total error between the data points and the fitted line or curve.

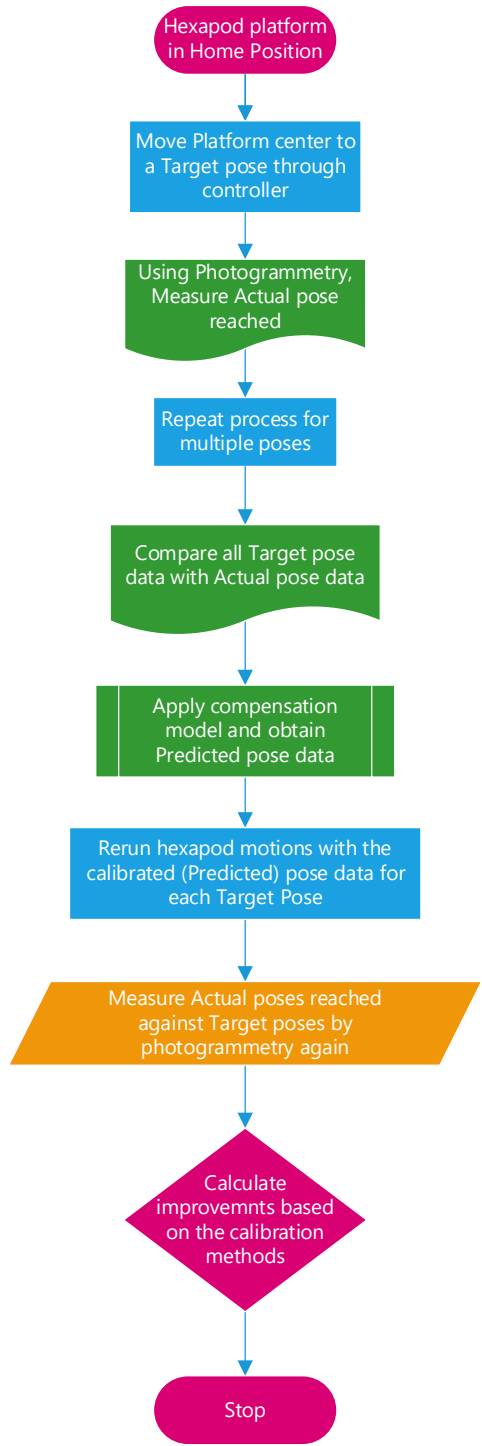

Figure 4: Flow chart for the adopted calibration process

Three different options were adopted by the authors to find the predicted poses after combining them with the correction values. These three strategies are explained in the next three sections.

The principle of the calibration process for all the three options has been explained in flow chat shown in Figure 4.

*1.2.1    Calibration – Option 1: With corrected DH parameters*

In this calibration approach, the Denavit-Hartenberg(DH) parameters for each pose have been computed through the new algorithm developed by the authors as explained in [15]. This algorithm enables one to find unique, feasible solutions for each platform pose by using forward kinematics. Employing this technique, the DH parameters in each actuator path for each target pose and corresponding actual pose were found. The errors for each pair of DH parameters were calculated. Using the least square method, the predicted DH parameters were calculated. The predicted DH parameters for each actuator path are used to find the predicted platform pose along each actuator path. For each target pose, there are 6 predicted poses calculated with compensated DH parameters through six actuator paths. Computing by matrix averaging technique, the 6 position vectors (through each actuator path) for each pose were unified and a new position vector for the predicted pose has been calculated. The matrix averaging of 6 orientation vectors from each actuator path for a pose did not yield any meaningful outcome, so they have not been treated in the same way as the position vectors. The orientation vectors for the predicted poses were left unchanged. A complete predicted pose vector is defined by combining the unified position vector with the target orientation vector (Euler angles) for that pose. (suffix '*uc*' has been used to indicate uncorrected or uncompensated values and suffix '*dc*' denotes the corrected or compensated values through DH parameters.)

If a target pose vector = $(x_{uc}\ y_{uc}\ z_{uc}\ \alpha_{uc}\ \beta_{uc}\ \gamma_{uc})^T$

and the predicted position vector after correction = $(x_{dc}\ y_{dc}\ z_{dc})^T$

then the new predicted pose for the target pose = $(x_{dc}\ y_{dc}\ z_{dc}\ \alpha_{uc}\ \beta_{uc}\ \gamma_{uc})^T$ (4)

*1.2.2 Calibration – Option 2: With corrected DH parameters & Euler angles*

In this option, the position vector for a pose has been corrected in the same way as described in the previous section.

However, the orientation vectors were treated differently. For each pose, there are 2 orientation vectors: one for the target pose and the other one for the actual measured pose. For *n* poses, *n* error values were calculated by subtracting the target orientations from the actual orientations. Using the Least-Square method, the compensation angles values for each pose have been calculated from error values. Now, each predicted pose has been calculated by creating a set of unified position vectors and compensated orientation vectors. Mathematically, it can be expressed as follows (suffix '*lc*' denotes least-square method compensated values):

Let, the target pose vector = $(x_{uc}\ y_{uc}\ z_{uc}\ \alpha_{uc}\ \beta_{uc}\ \gamma_{uc})^T$

The predicted translation vector after correction = $(x_{dc}\ y_{dc}\ z_{dc})^T$

The predicted orientation vector after correction as stated above = = $(\alpha_{lc}\ \beta_{lc}\ \gamma_{lc})^T$

then the new predicted pose for the target pose = $(x_{dc}\ y_{dc}\ z_{dc}\ \alpha_{lc}\ \beta_{lc}\ \gamma_{lc})^T$ (5)

*1.2.3 Calibration – Option 3: With corrected translation vectors & Euler angles*

In this option the predicted pose for a target pose has been calculated by correcting both position vectors and orientation vectors through the Least-Square method.

For *n* number of poses there are *n* numbers of target poses and actual poses. Each pose has 6 parameters, these 6 parameters are a combination of a position vector and an orientation vector.

For *n* numbers of poses, *n* error values were calculated by subtracting the target pose parameters from the actual pose parameters. Using the Least-Square method, the compensation values for each pose parameter have been calculated from error values and applied to the target poses. These compensated pose parameters are used as predicted poses for each target pose. The main difference between this option and the earlier two options is that in this case the DH parameters were not considered for finding the predicted poses. The new vector calculation can be expressed in the following way:

Let, the target pose vector = $(x_{uc} \ y_{uc} \ z_{uc} \ \alpha_{uc} \ \beta_{uc} \ \gamma_{uc})^T$

The predicted position vector after correction = $(x_{lc} \ y_{lc} \ z_{lc})^T$

The predicted orientation vector after correction = $(\alpha_{lc} \ \beta_{lc} \ \gamma_{lc})^T$

then the new predicted pose for the target pose = $(x_{lc} \ y_{lc} \ z_{lc} \ \alpha_{lc} \ \beta_{lc} \ \gamma_{lc})^T$ (6)

The above three options can be summarized in the Table 1

Table 1: Summery of the experiment options

| Option no. | Position vector correction | Orientation vector correction |
|---|---|---|
| 1 | Through DH parameters | No correction |
| 2 | Through DH parameters | Least Square method |
| 3 | Least Square method | Least Square method |

After position and orientation error ranges were calculated for uncompensated and compensated poses, the magnitude of errors for position and orientation have been calculated using the "Root Mean Square Error" (RMSE) equation:

$$\text{Magnitude of error} = \sqrt{\frac{(x\_range^2 + y\_range^2 + z\_range^2)}{3}} \tag{7}$$

## 3. Experimental Setup & Data Collection

The experimental setup for these calibration processes did not need any special hardware beyond one high resolution digital camera. This camera is independent from the hexapod test fixture and does not interfere with the operations of the system. A Nikon D3200 digital camera has been used for this purpose. The lens used is Nikon AF-S DX Nikkor 18-105mm. The camera has been used in "manual" mode. This was a requirement by the photogrammetry software to keep the camera calibration and other image parameters uniform throughout the process. The camera settings were kept fixed once the camera calibration was done. Though the zoom value used in the process is not visible on the camera setting windows shown in the Figure 5, the zoom settings are also kept fixed to maintain the same values for the lens parameters.

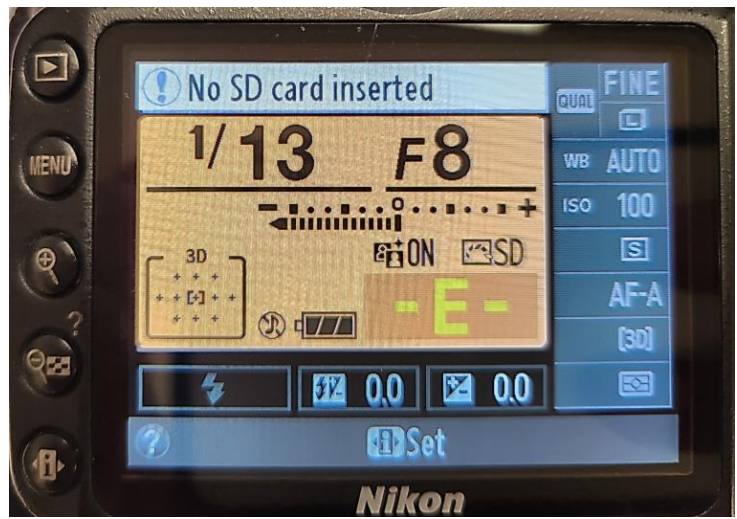

Figure 5: Nikon D3200 camera settings for Photogrammetry

In this experimental setup, the initial step involved the calibration of the camera in accordance with the specific calibration procedure prescribed by the photogrammetry software, "Photomodeler". To accomplish this, a series of printed templates, shown in Figure 6, were employed. Subsequently, an automated camera calibration process was executed once these template images were processed through Photomodeler. Upon the successful completion of this calibration procedure, the camera was ready for the project. The completed calibration data was recorded and stored in a file, which was subsequently referenced during the image processing stage for the calibration of Tiger 66.1.

In the next step, 34 random poses were selected in the moving gripper's workspace. These workspaces were free from "singularity" condition and that has been verified in MATLAB code before using in the control software. The mathematical check for singularity is included inside the calculation code by checking if the determinant of the force Jacobian matrix in that pose is zero or not [20]. For each pose three images were taken in three different angles to satisfy the need for photogrammetric measurements. These three images were processed in Photomodeler as shown in Figure 7. Each image is processed manually, and they were integrated to generate a 3D wire frame model (visible in the extreme right-side window of the Figure 7).

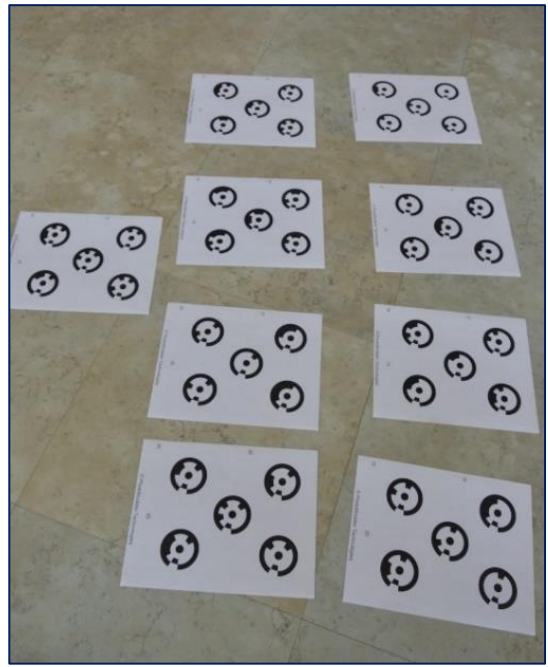

Figure 6: Calibration templates

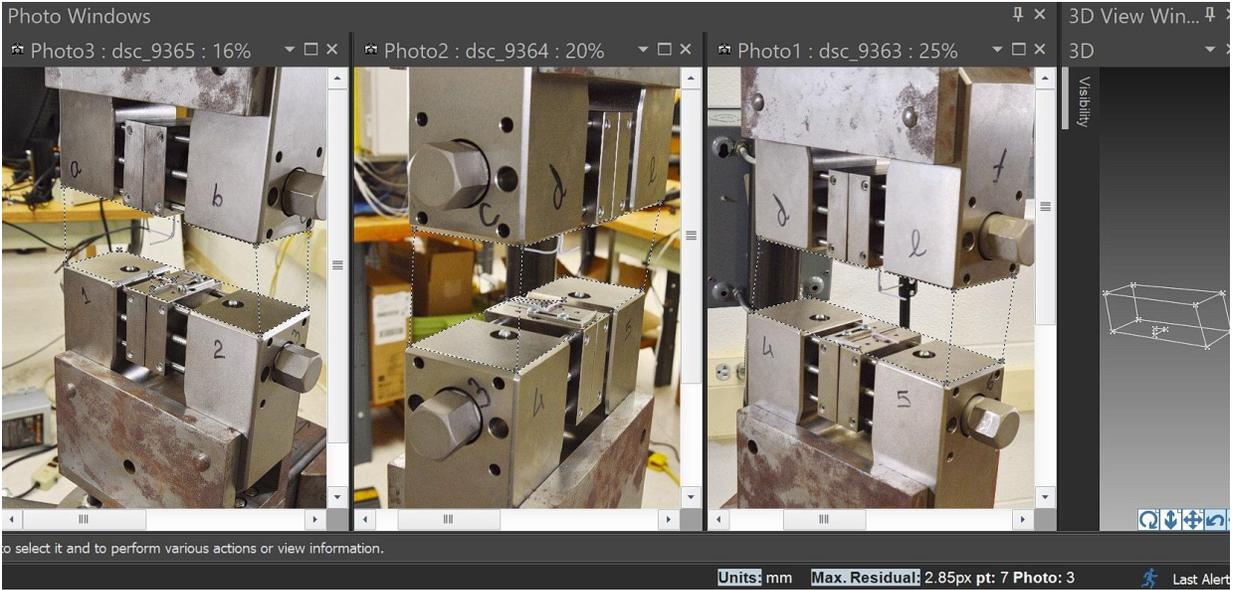

Figure 7: Photomodeler user interface for image processing

This process is repeated for all poses before and after calibrations following 3 options described in section 2.

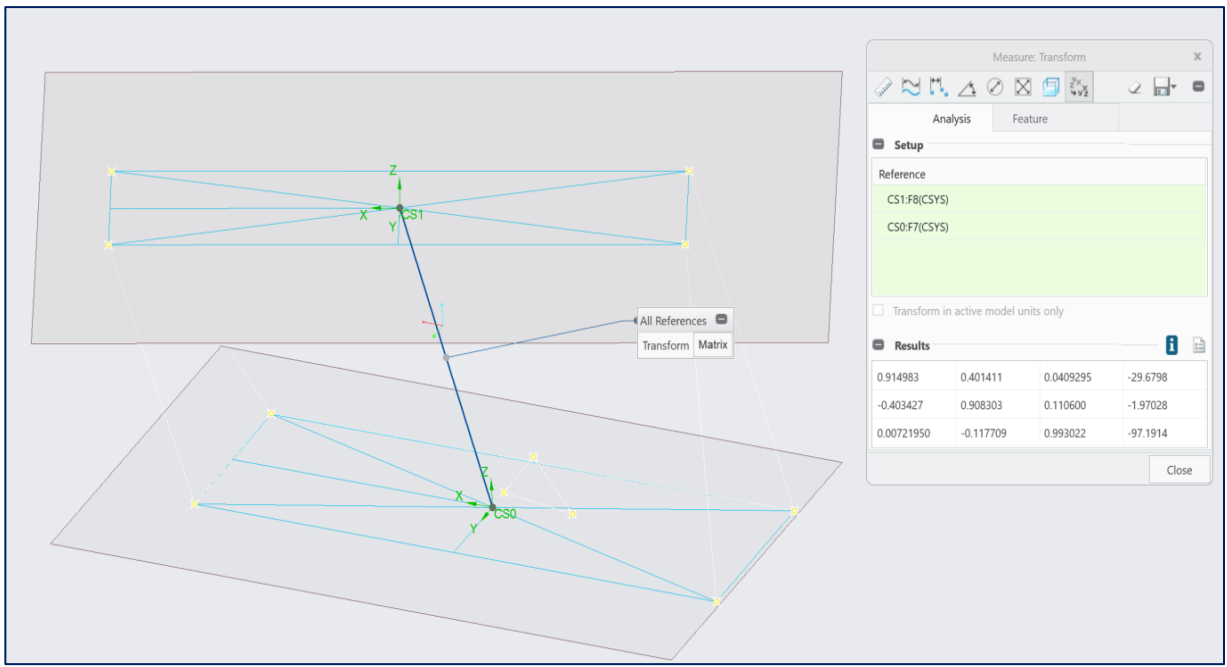

Figure 8: Manual processing of wire frame model in ProEngineer Creo

The wire frame model now becomes the input for the ProEngineer Creo software. Each wire frame model is manually analyzed to collect the pose data. A typical screenshot from Creo is shown in Figure 8 after the complete processing of a wire frame model in this software package. The outcome from this process is the translation vector and rotation matrix for the actual motion performed by the hexapod. The 3 rows of the last column in the 3x4 matrix shown in Figure 8 denotes the position vector and the remaining 3x3 matrix denotes the orientation matrix. By using MATLAB function, the rotation values about the axes are extracted from this 3x3 matrix.

## 4. Results and Analysis

The 6 parameters for all the poses were extracted by photogrammetry and recorded. In the beginning the target poses and actual poses were compared by measuring the differences between the respective pose parameters. The absolute difference values are plotted and shown in Figure 9. A random 34 target poses in the workspace were considered for the experiment and the lower grip center has been moved to those poses one by one. The actual poses reached by the lower grip center in this process are uncalibrated poses and obtained from as-build condition. The prefix '-*tran*' has been used to denote the position variables and '-*rot*' has been used to denote the orientation variables.

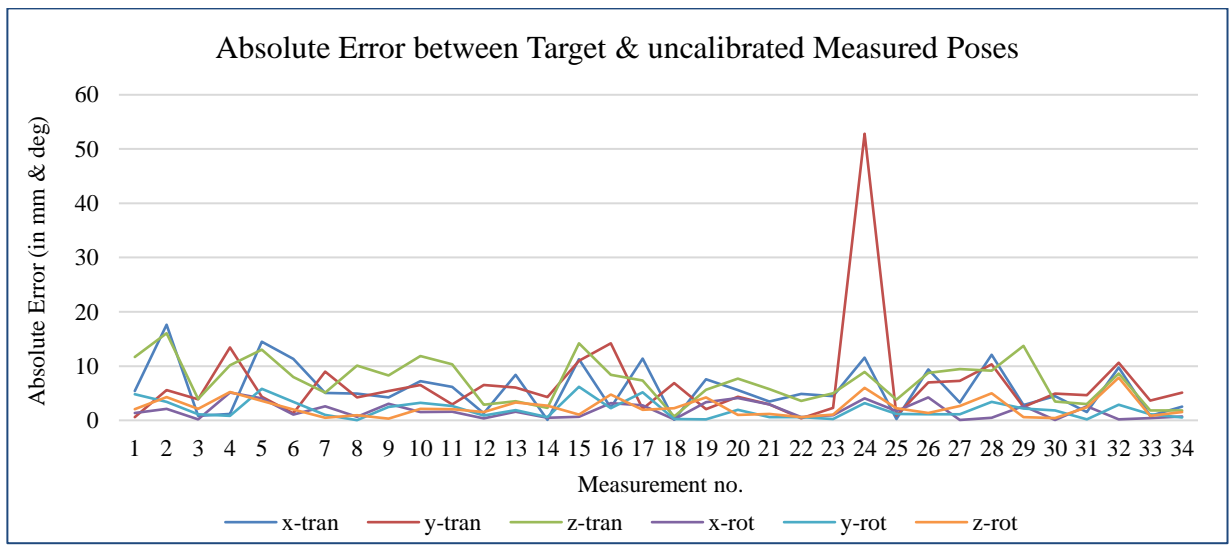

Figure 9: Absolute Error between Target & uncalibrated Measured poses

From the chart it is seen that pose number 24 has one parameter which is significantly out of range compared to the other values. It was determined that while this point is not singular, it is close to a singular point. Thus, it was designated as an outlier and removed from future calculations. The error range of the uncalibrated poses are shown in Figure 10.

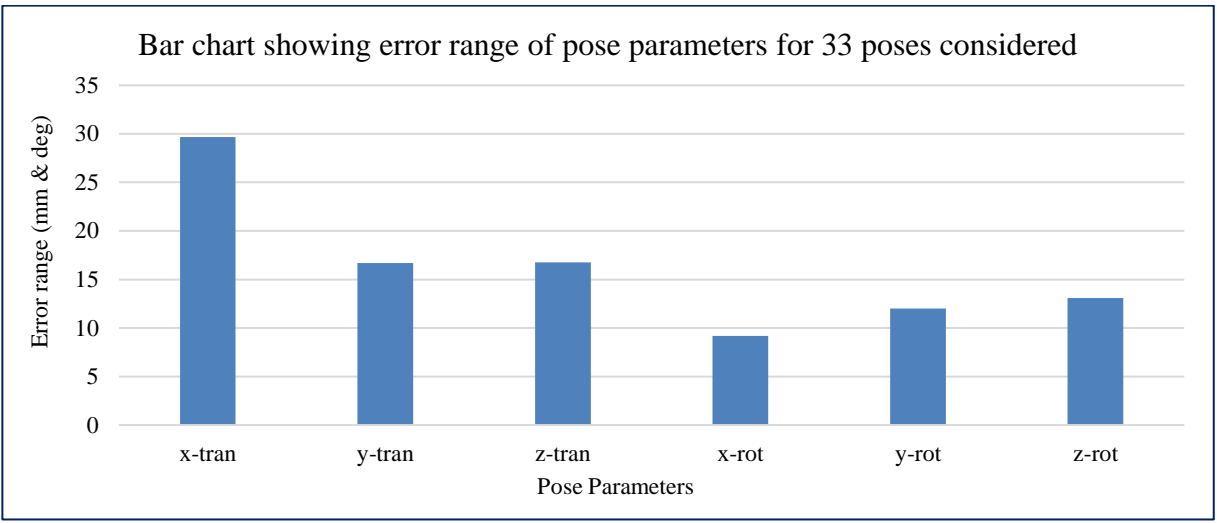

Figure 10: Error range of the pose parameters in uncalibrated condition

*1.4.1 Calibration results: Option 1*

Calibrating as per options, the lower grip center has been moved to the new predicted positions for the corresponding target position. During this calibration process, some of the predicted pose parameter values after compensating fell outside of the hexapod operating range and they were removed from further calculations and analysis. The total valid pose numbers came down to 27 after removing the out-of-range values. The hexapod has been instructed by the controller to move to new predicted poses one by one. The actual pose measurements were done through photogrammetry and pose data were compared with the corresponding target pose data. Bar chart in Figure 11 shows the error range of 6 pose parameters before and after calibration.

There are improvements in pose parameters except position values along y and z axis. For all other parameters various measures of improvement were observed.

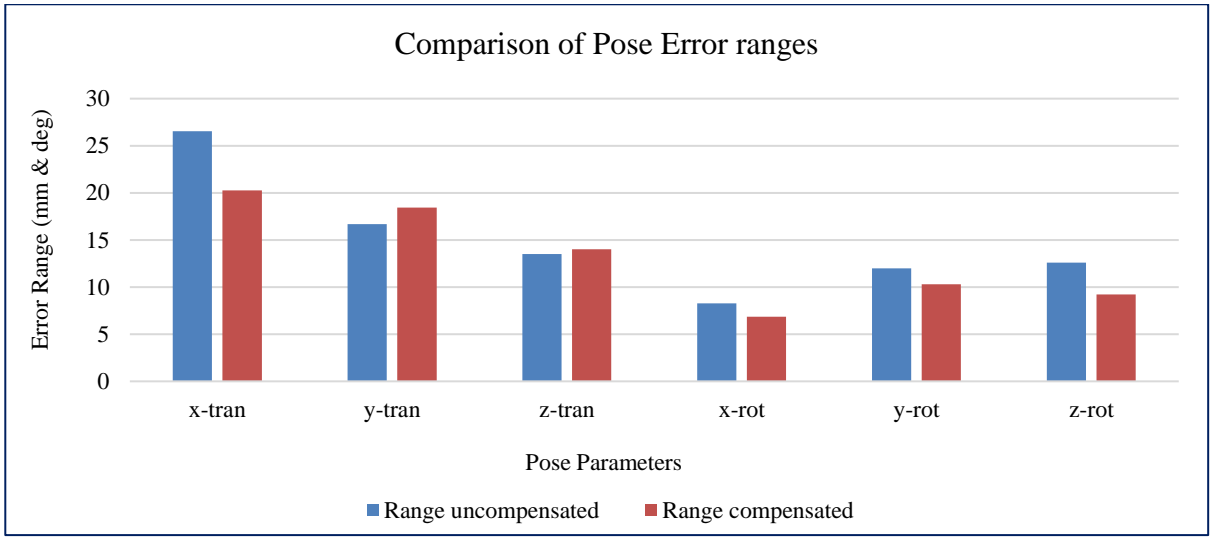

Figure 11: Comparison of pose parameters error range after calibration as per Option 1

The absolute deviations of each pose parameters before and after the calibration are shown in Figure 12. Here the suffix '*c*' with axis identifier denotes the calibrated values.

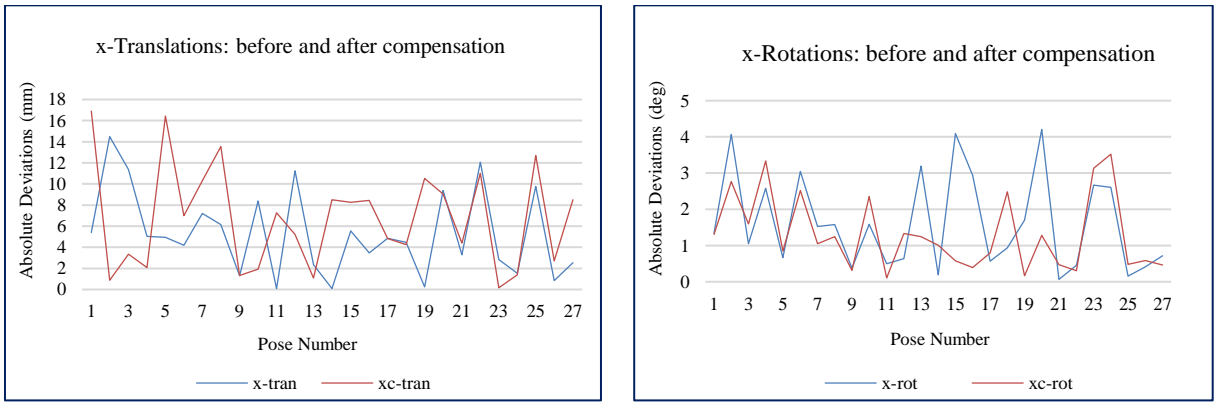

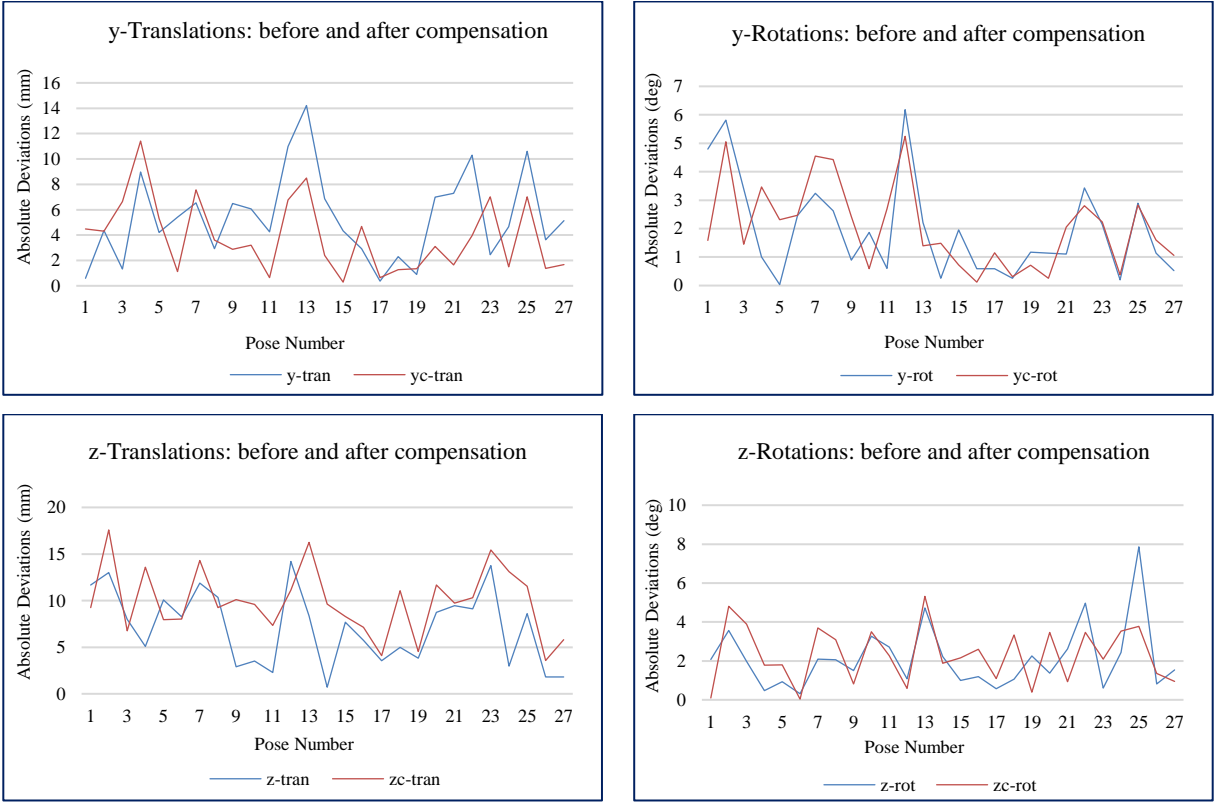

Figure 12: Comparison of absolute deviations of all 6 pose parameters for option 1

To better understand the impact of the calibration method used as per option 1, the Magnitude of errors for position (measured in mm) and orientation parameters (measured in deg) were compared. The data were shown in the Table 2. The magnitude of errors for position and orientation values were calculated separately. The results show that the magnitude of error for position improved by 9.8% and for orientation improvement is 19.9%.

Table 2: Pose improvement after calibration as per option 1

| **Parameters** | **x-tran** | **y-tran** | **z-tran** | **x-rot** | **y-rot** | **z-rot** |
|---|---|---|---|---|---|---|
| Error range-uncompensated | 26.54 | 16.67 | 13.49 | 8.27 | 11.99 | 12.59 |
| Magnitude of errors - uncompensated | | 19.70 | | | 11.12 | |
| Error range-compensated | 20.25 | 18.44 | 14.01 | 6.86 | 10.29 | 9.2 |
| **Magnitude of errors - compensated** | | 17.76 | | | 8.90 | |
| **Magnitude of improvement %** | | **9.8%** | | | **19.9%** | |

## 1.4.2 Calibration results: Option 2

In this calibration method, the parameter values for some of the predicted poses were out of the hexapod's travel range after compensation. Those poses were discarded from considerations. After removing those poses, calibration calculations were done with 26 poses. The error range of the uncalibrated and calibrated pose parameters are shown in Figure 13.

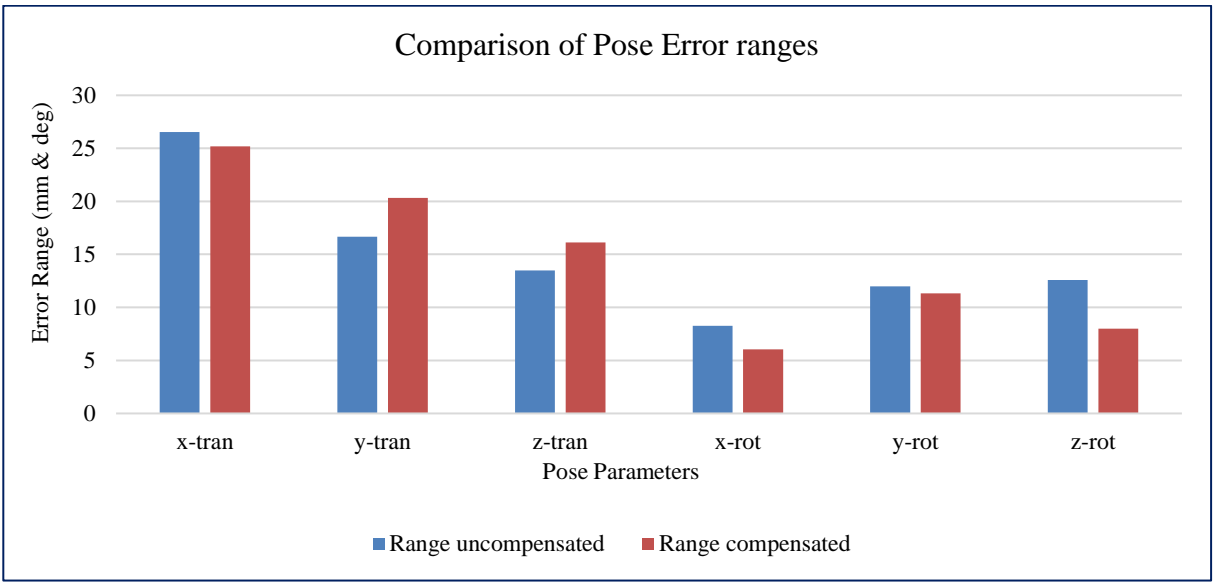

Figure 13: Comparison of pose parameters error range after calibration as per Option 2

The individual parameters deviations comparison can be seen in the Figure 14.

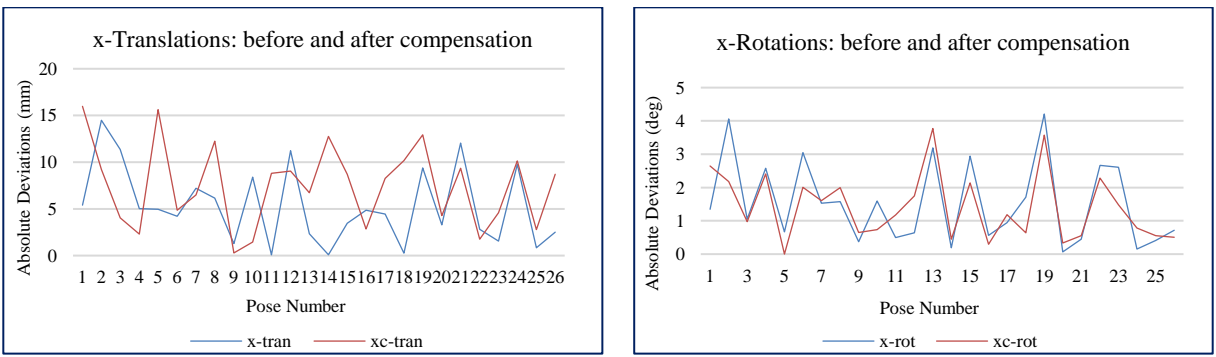

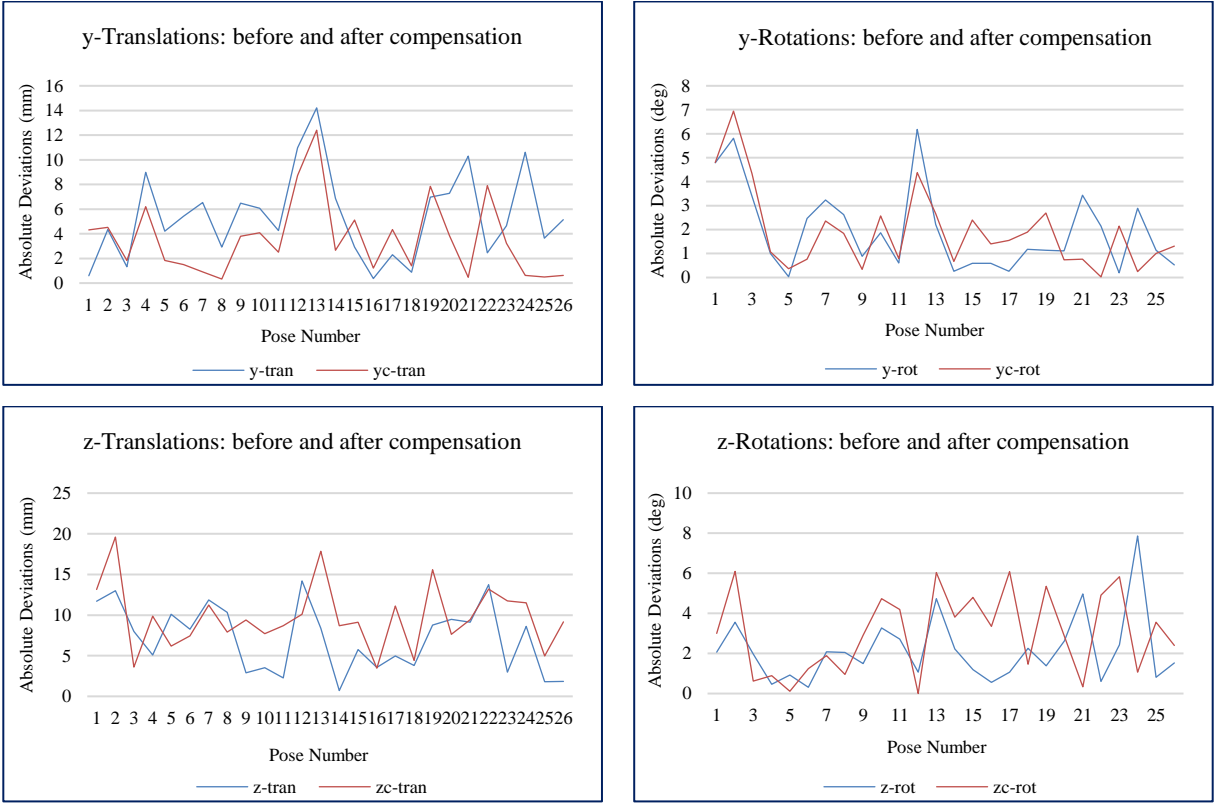

Figure 14: Comparison of absolute deviations of all 6 pose parameters for option 2

The pose parameters improvement based on this method of calibration is calculated and shown in Table 3.

Table 3: Pose improvement after calibration as per option 2

| Parameters | x-tran | y-tran | z-tran | x-rot | y-rot | z-rot |
|---|---|---|---|---|---|---|
| Error range -uncompensated | 26.54 | 16.67 | 13.49 | 8.27 | 11.99 | 12.59 |
| Magnitude of errors - uncompensated | | 19.70 | | | 11.12 | |
| Error range -compensated | 25.20 | 20.31 | 16.14 | 6.05 | 11.32 | 7.99 |
| Magnitude of errors - compensated | | 20.88 | | | 8.73 | |
| **Magnitude of improvement %** | | **-6.0%** | | | **21.5%** | |

In this case, the magnitude of error for the positional part of the pose deteriorated by 6%, though there is 21.5% improvement observed in the orientation part.

## 1.4.3 Calibration results: Option 3

In the earlier two options, the position parameters were corrected using DH parameter corrections through the Least-Square methods. In this option, both position and orientation parameters were compensated by using only Least-Square methods from the error values obtained from the uncompensated measurements and comparing those values with the target values. After compensating the pose parameters in this method, only 21 poses were found to be inside the valid workspace of the lower grip center of Tiger 66.1. The comparison of the error ranges before and after calibration is shown in the Figure 15.

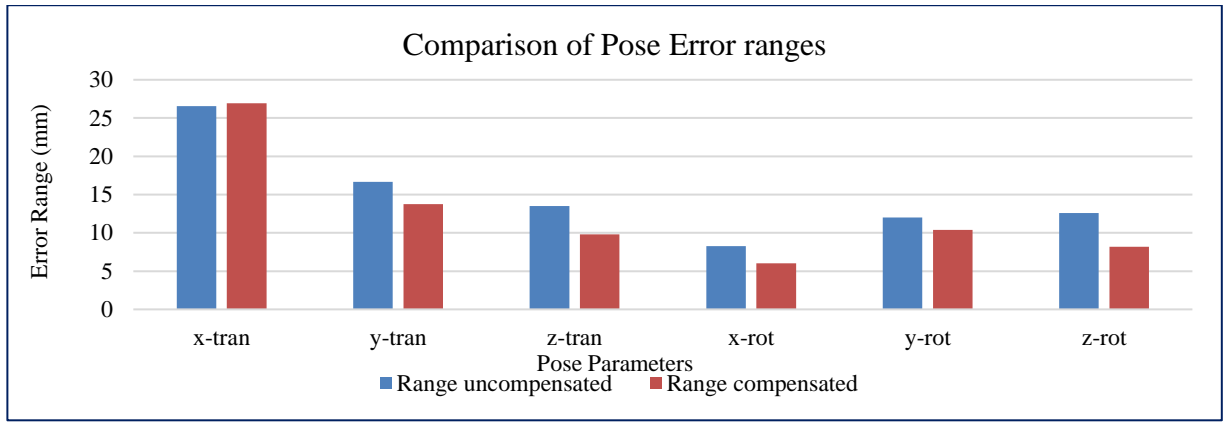

Figure 15: Comparison of pose parameters error range after calibration as per Option 3

The comparison of deviation of each uncompensated and compensated pose parameters with respect to the target pose has been presented in the Figure 16.

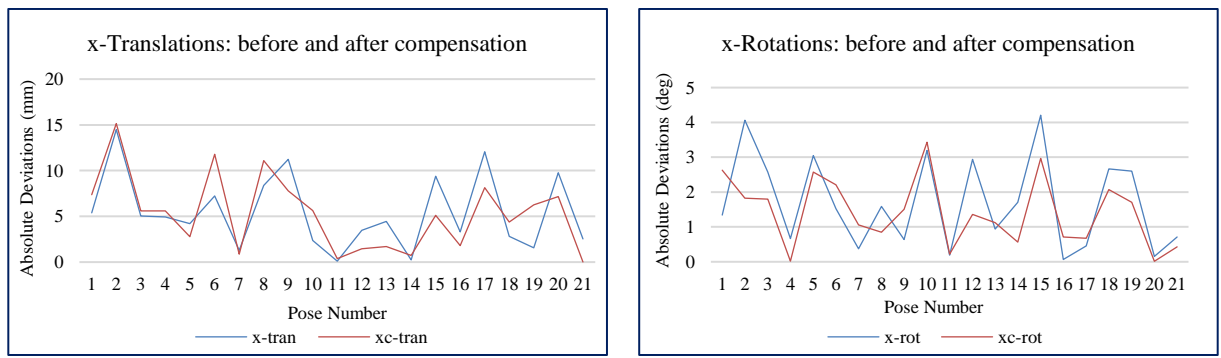

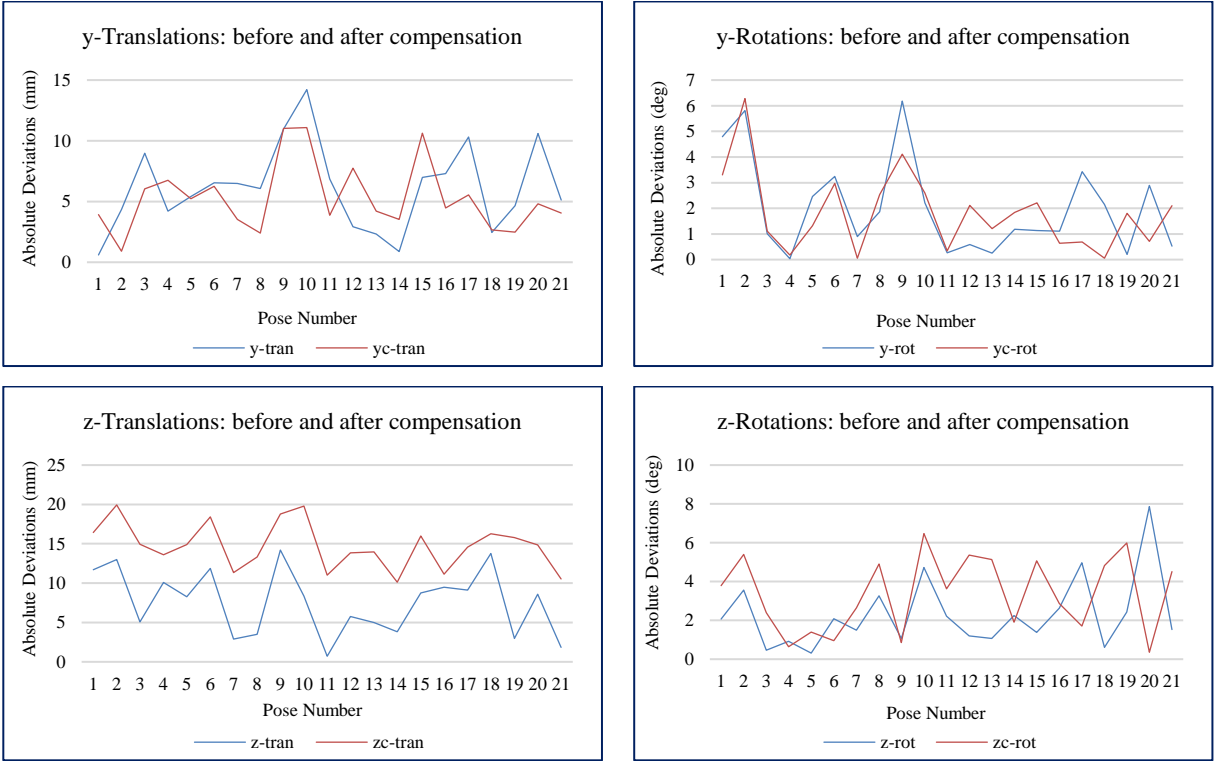

Figure 16: Comparison of absolute deviations of all 6 pose parameters for option 3

The comparison of magnitude of error has been captured in Table 4.

Table 4: Pose improvement after calibration as per option 3

| Parameters | x-tran | y-tran | z-tran | x-rot | y-rot | z-rot |
|---|---|---|---|---|---|---|
| Error range-uncompensated | 26.54 | 16.67 | 13.49 | 8.27 | 11.99 | 12.59 |
| Magnitude of errors - uncompensated | | 19.70 | | | 11.12 | |
| Error range-compensated | 26.92 | 13.74 | 9.81 | 6.01 | 10.39 | 8.18 |
| Magnitude of errors - compensated | | 18.35 | | | 8.39 | |
| **Magnitude of improvement %** | | **6.9%** | | | **24.6%** | |

The magnitude of error for both position and orientation observed to be improved and their values are 6.9% and 24.6% respectively.

## 5. Discussions

The hexapod test-frame Tiger 66.1 has been subjected to operation and calibration for the first time after its fabrication. For the first time calibration process, very basic methods have been planned to start with. The Least-Square method is one of the basic statistical methods for fitting a line or curve to a set of data points to minimize errors on the data points. The above-mentioned three options have been considered as the starting point before moving to further complex compensation methods.

In option 1, the error minimization was done by finding the DH parameters for each pose. But it has been observed during the calculation that the same calculation methods cannot be applied for angular vectors. Therefore, the position and orientation vectors of a pose have been treated differently due to the units involved. The target orientation vector used for the calibrated predicted poses without any change. In the second option the same strategy as option 1 has been followed for position and orientation vectors were compensated by least-square methods. And in the third option, the error corrections for both the vectors were done with least-square methods taking the pose parameter differences into consideration.

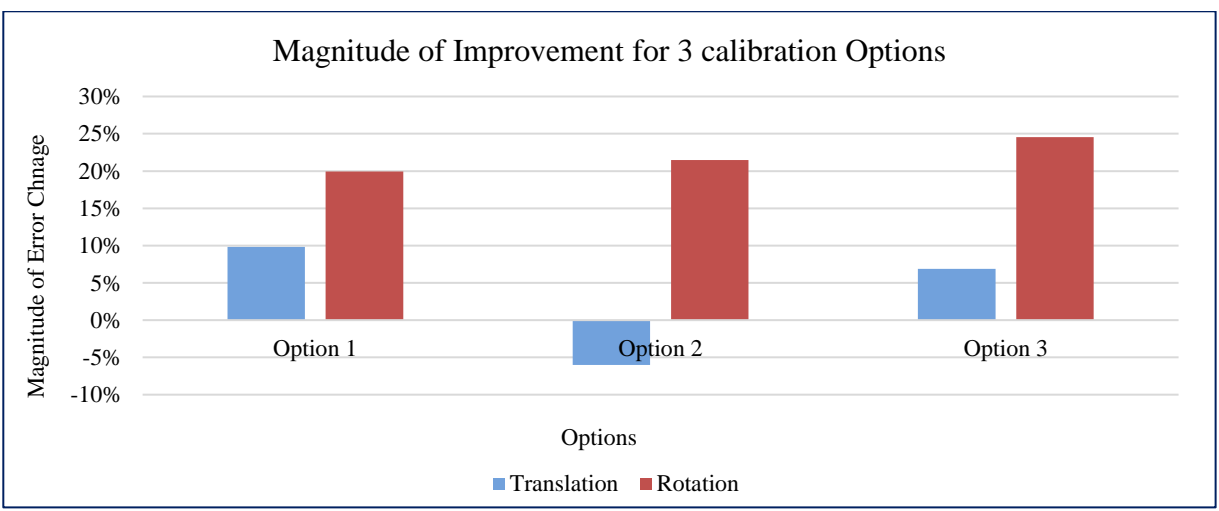

Figure 17: Comparison of Magnitude of Improvement for all 3 Calibration strategies

The three calibration strategies yielded different results and different amounts of pose-accuracy improvements. The magnitude of improvements for all three options were calculated and presented together Figure 17. As can be seen from the plot, option 2 shows a reduction in positional accuracy, whereas option 1 & 3 have shown improvements in both position and orientation accuracy. In option 1 the improvement in both position and orientation accuracy appears to be more balanced (the measure of improvement percentages is closer) than option 3.

When the error ranges are compared between the uncompensated pose values and compensated pose values from all three options, option 1 shows more steady changes than other two options; though the error ranges for y-position and z-position increased marginally after the calibration process based on this method. Option 3 has shown maximum error reduction except for x-position values. The error range comparisons for these options were shown in Figure 18.

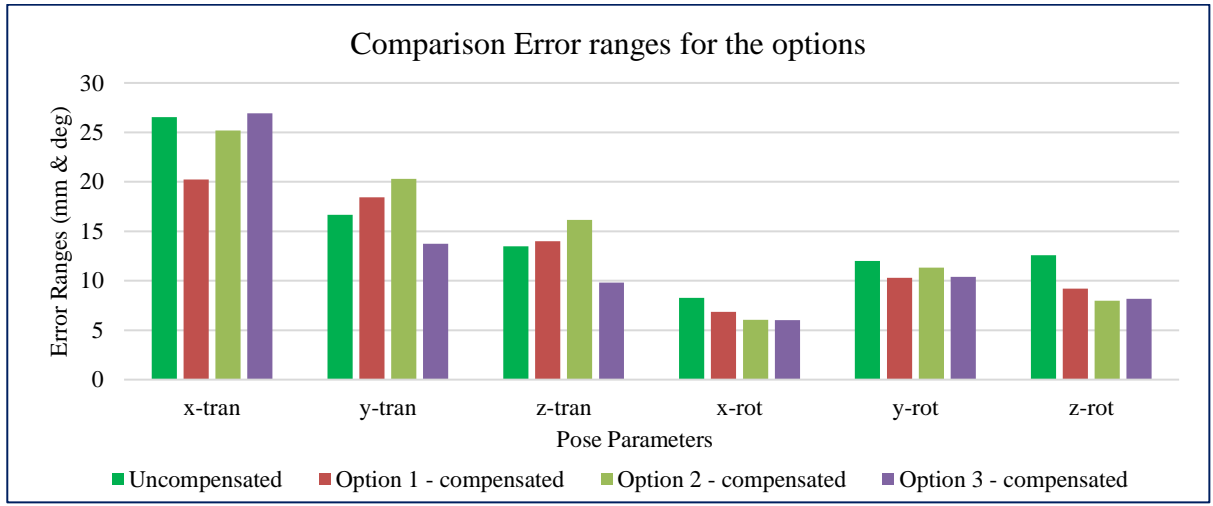

Figure 18: Comparison of error ranges for all options

In this research, various calibration methods were diligently applied to enhance the accuracy of hexapod poses. Despite the improvements in the hexapod poses following each of these calibration methods, none of them display a clear indication that which one is best and thereby which method is best suited for this system. It is crucial to acknowledge that there are multiple sources of errors which affect the pose accuracy of a hexapod. For this hexapod, those errors factors might have more complex relationships that is impacting the pose accuracy, the Least Square method is inadequate to comprehensively capturing the complex interdependencies among multiple error factors influencing overall pose error. Identifying a more robust relationship from error data could serve as a focus for future research endeavor. These calibration methods were tried on Tiger 66.1 for the first time after it became operational. The control software has also been developed and used with its first version. While definitive conclusions regarding the optimal calibration method remain elusive, it is evident that the hexapod exhibited notable improvements in response to initial calibration processes. Among the strategies explored by the authors, option 1 appears to be a more robust calibration model. If there are any constructional errors in Tiger 66.1, those errors can be taken care of with correction of the DH parameters without going into more detailed measurement of the fabrication errors.

Another important factor to be considered in this experiment is the number of poses considered. The experiment started with 34 random poses, and it came down to 21 poses. The efficacy of compensation models relies on the number of data points, it is expected that considering more numbers of poses for the experiments may help to further refine the calibration results. During the process some of the pose points had to be eliminated from the calculations due to the new predicted poses lying outside the workspace and out of range of the hexapod's motion. Consequently, a greater number of initial random poses could increase the number of compensated valid poses that can be considered for calibration calculation and analysis. Furthermore, there is likely considerable merit in considering the poses most likely to be used, i.e. those that would lie along the anticipated load pathways (tension, compression, torsion, bending, and combinations thereof).

It is imperative to acknowledge the potential contribution of software and image processing errors to the calculated results. However, both software platforms employed in these calibration techniques are well-established and widely adopted within industrial contexts, thereby minimizing the likelihood of software-related errors. Any residual errors can primarily be attributed to manual processing estimation. However, for image processing in this research, the error limit was restricted to below 5 pixels which is equal to 0.425mm (for 300dpi resolution 1pixel = 0.085mm), thereby ensuring minimal impact on the overall accuracy of the calculations.

## 6. Conclusion

The target of this project was to estimate pose errors in Tiger 66.1 using a non-invasive, least instrumented approach while exploring simple calibration methods to enhance hexapod pose accuracy. The implementation of photogrammetry for pose measurements is a successful achievement in this effort. It shows promise for further development of photogrammetry method though the initial iteration yielded modest improvements. There are several benefits of using the photogrammetry method: it requires minimal modifications to the primary system hardware, and as no additional sensors are needed this is a cost effective and time efficient method. However, challenges may arise in certain situations, particularly when space constraints hinder the installation of the camera setup around the system. Additionally, achieving superior measurement accuracy may necessitate expensive camera systems, while advanced image processing demands high-power hardware and software. These challenges present limitations in the practical implementation of photogrammetry methods in all situations.

The initial controller software efficiently manages the hexapod and incorporates compensations from the three calibration methods employed. Although the accuracy gains are not substantial, these methods demonstrate the possibility of employing non-invasive photogrammetry for hexapod calibration. One of the methods used by the authors involved forward kinematics to derive a unique feasible solution and calculate DH parameters, and the Least-Squares method displayed some error reduction potential, suggesting at the possibility of investigating more complex error compensation models. This successful calibration through photogrammetry not only enhances the hexapod's overall performance but also opens avenues for its versatile deployment across various domains, ranging from industrial automation to advanced research initiatives. This simple, effective calibration process is a significant step toward achieving high quality measurements in diverse applications, thereby contributing substantially to the progress of automation and robotics. In the next phase of this research, this non-invasive photogrammetry method is favorable to use to calibrate Tiger 66.1 under complex loading condition. The main advantage is that there is no need for additional instrumentation. The potential for continued refinement and innovation in this field is vast, with photogrammetry emerging as a valuable tool in enhancing the accuracy and reliability of hexapod systems and, by extension, a wide spectrum of robotic and automated processes through non-invasive and minimal instrumented methods.

**FUNDING**

The authors express their appreciation to Clemson University who financially supported this work and to the United States Naval Research Laboratory in Washington, D.C. who provided technical insight and support for the design and operation of the TIGER 66.1 system through NCRADA-NRL-20-719.

**COMPETING INTERESTS**

The authors have no relevant financial or non-financial interests to disclose.

## AUTHOR CONTRIBUTIONS

All authors contributed to the conception and design of the research. Software coding, data collection and analysis, and preparation of the first draft were performed by Sourabh Karmakar. All authors commented on and contributed to previous manuscript versions. All authors read and approved the final manuscript.

## REFERENCES

[1] B. Dasgupta and T. S. Mruthyunjaya, "A constructive predictor-corrector algorithm for the direct position kinematics problem for a general 6-6 Stewart platform," *Mech. Mach. Theory*, vol. 31, no. 6, pp. 799–811, 1996, doi: 10.1016/0094-114X(95)00106-9.

[2] W. Tanaka, T. Arai, K. Inoue, Y. Mae, and C. S. Park, "Simplified kinematic calibration for a class of parallel mechanism," *Proc. - IEEE Int. Conf. Robot. Autom.*, vol. 1, no. May, pp. 483–488, 2002, doi: 10.1109/ROBOT.2002.1013406.

[3] D. Daney, I. Z. Emiris, Y. Papegay, E. Tsigaridas, and J. P. Merlet, "Calibration of parallel robots : on the Elimination of Pose-Dependent Parameters," *EuCoMeS 2006 - 1st Eur. Conf. Mech. Sci. Conf. Proc.*, pp. 1–12, 2006.

[4] A. C. Majarena, J. Santolaria, D. Samper, and J. J. Aguilar, "An overview of kinematic and calibration models using internal/external sensors or constraints to improve the behavior of spatial parallel mechanisms," *Sensors (Switzerland)*, vol. 10, no. 11, pp. 10256–10297, 2010, doi: 10.3390/s101110256.

[5] S. Karmakar and C. J. Turner, "A Literature Review on Stewart – Gough Platform Calibrations," vol. 146, no. August, pp. 1–12, 2024, doi: 10.1115/1.4064487.

[6] D. Jakobovi, L. Budin, D. Jakobović, and L. Budin, "Forward kinematics of a Stewart platform mechanism," *J. Mech. Des.*, vol. 115, no. 4, pp. 277–282, 2002, [Online]. Available: https://bib.irb.hr/datoteka/89476.ines2002.pdf%0Ahttp://citeseerx.ist.psu.edu/viewdoc/download?doi=10.1.1.13.580&rep=rep1&type=pdf

[7] H. Zhuang, J. Yan, and O. Masory, "Calibration of Stewart platforms and other parallel manipulators by minimizing inverse kinematic residuals," *J. Robot. Syst.*, vol. 15, no. 7, pp. 395–405, 1998, doi: 10.1002/(SICI)1097-4563(199807)15:7<395::AID-ROB2>3.0.CO;2-H.

[8] J. Ryu and A. Rauf, "A new method for fully autonomous calibration of parallel manipulators using a constraint link," *IEEE/ASME Int. Conf. Adv. Intell. Mechatronics, AIM*, vol. 1, no. July, pp. 141–146, 2001, doi: 10.1109/aim.2001.936444.

[9] K. Großmann, B. Kauschinger, and S. Szatmári, "Kinematic calibration of a hexapod of simple design," *Prod. Eng.*, vol. 2, no. 3, pp. 317–325, 2008, doi: 10.1007/s11740-008-0092-6.

[10] Y. Liu, B. Liang, C. Li, L. Xue, S. Hu, and Y. Jiang, "Calibration of a Steward parallel robot using genetic algorithm," *Proc. 2007 IEEE Int. Conf. Mechatronics Autom. ICMA 2007*, pp. 2495–2500, 2007, doi: 10.1109/ICMA.2007.4303948.

[11] D. Daney, Y. Papegay, and B. Madeline, "Choosing measurement poses for robot calibration with the local convergence method and Tabu search," *Int. J. Rob. Res.*, vol. 24, no. 6, pp. 501–518, 2005, doi: 10.1177/0278364905053185.

[12] T. Dallej, H. Hadj-Abdelkader, N. Andreff, and P. Martinet, "Kinematic Calibration of a Gough-Stewart platform using an omnidirectional camera," *IEEE Int. Conf. Intell. Robot. Syst.*, pp. 4666–4671, 2006, doi: 10.1109/IROS.2006.282253.

[13] M. J. Nategh and M. M. Agheli, "A total solution to kinematic calibration of hexapod machine tools with a minimum number of measurement configurations and superior accuracies," *Int. J. Mach. Tools Manuf.*, vol. 49, no. 15, pp. 1155–1164, 2009, doi: 10.1016/j.ijmachtools.2009.08.009.

[14] S. Kucuk and Z. Bingul, *Robot Kinematics: Forward and Inverse Kinematics*, no. December. 2006. doi: 10.5772/5015.

[15] S. Karmakar and C. J. Turner, "Forward kinematics solution for a general Stewart platform through iteration based simulation," *Int. J. Adv. Manuf. Technol.*, no. 0123456789, 2023, doi: 10.1007/s00170-023-11130-9.

[16] S. T. Fry and C. J. Turner, "Design of a Stewart-Gough Platform for Engineering Material Characterization," *Proc. ASME 2016 Int. Mech. Eng. Congr. Expo.*, pp. 1–9, 2016, doi: 10.1115/imece2016-66669.

[17] Y. K. Yiu, J. Meng, and Z. X. Li, "Auto-calibration for a parallel manipulator with sensor redundancy," *Proc. - IEEE Int. Conf. Robot. Autom.*, vol. 3, pp. 3660–3665, 2003, doi: 10.1109/robot.2003.1242158.

[18] J. Michaloski, "Coordinated Joint Motion Control for an Industrial Robot," *NIST Interagency/Internal Rep. (NISTIR), Natl. Inst. Stand. Technol. Gaithersburg, MD,* no. March, 1988, [Online]. Available: https://tsapps.nist.gov/publication/get_pdf.cfm?pub_id=820236

[19] Wikipedia, "Least-Square method." Wikimedia Foundation, 2023. [Online]. Available: en.wikipedia.org/wiki/Least_squares_method.

[20] B. Li, Y. Cao, Q. Zhang, and Z. Huang, "Position-singularity analysis of a special class of the Stewart parallel mechanisms with two dissimilar semi-symmetrical hexagons," *Robotica*, vol. 31, no. 1, pp. 123–136, 2013, doi: 10.1017/S0263574712000148.